\documentclass[11pt]{article}

\usepackage[margin=1in]{geometry}
\usepackage[utf8]{inputenc}

\usepackage{booktabs}
\usepackage{tabularx}
\usepackage{multirow}
\usepackage{array}

\usepackage{graphicx}
\graphicspath{{figures/}}
\usepackage{float}

\usepackage{natbib}
\usepackage[colorlinks=true,linkcolor=blue,citecolor=blue,urlcolor=blue]{hyperref}

\usepackage{enumitem}

\usepackage{amsmath}
\usepackage{amssymb}

\title{Multimodal Deep Learning for Early Prediction of Patient Deterioration in the ICU:\\ Integrating Time-Series EHR Data with Clinical Notes}
\author{Binesh Sadanandan}
\date{}

\begin{document}

\maketitle

\begin{abstract}
Early identification of patients at risk for clinical deterioration in the intensive care unit (ICU) remains a critical challenge. Delayed recognition of impending adverse events, including mortality, vasopressor initiation, and mechanical ventilation, contributes to preventable morbidity and mortality. We present a multimodal deep learning approach that combines structured time-series data (vital signs and laboratory values) with unstructured clinical notes to predict patient deterioration within 24 hours. Using the MIMIC-IV database, we constructed a cohort of 74,822 ICU stays and generated 5.7 million hourly prediction samples. Our architecture employs a bidirectional LSTM encoder for temporal patterns in physiologic data and ClinicalBERT embeddings for clinical notes, fused through a cross-modal attention mechanism. We also present a systematic review of existing approaches to ICU deterioration prediction, identifying 31 studies published between 2015 and 2024. Most existing models rely solely on structured data and achieve area under the curve (AUC) values between 0.70 and 0.85. Studies incorporating clinical notes remain rare but show promise for capturing information not present in structured fields. Our multimodal model achieves a test AUROC of 0.7857 and AUPRC of 0.1908 on 823,641 held-out samples, with a validation-to-test gap of only 0.6 percentage points. Ablation analysis validates the multimodal approach: clinical notes improve AUROC by 2.5 percentage points and AUPRC by 39.2\% relative to a structured-only baseline, while deep learning models consistently outperform classical baselines (XGBoost AUROC: 0.7486, logistic regression: 0.7171). This work contributes both a thorough review of the field and a reproducible multimodal framework for clinical deterioration prediction.
\end{abstract}

\section{Introduction}

The intensive care unit represents one of the most data-rich environments in modern healthcare. Patients are continuously monitored, generating thousands of physiologic measurements per day. Laboratory values, medication administrations, and clinical assessments accumulate in electronic health records. Yet despite this wealth of information, recognizing early signs of clinical deterioration remains challenging \citep{Churpek2016}. Too often, adverse events like cardiac arrest, respiratory failure, or septic shock are identified only after they've progressed to the point where intervention options become limited \citep{Hillman2005}.

Early warning scores have attempted to address this problem. Systems like NEWS (National Early Warning Score) and MEWS (Modified Early Warning Score) use threshold-based rules on vital signs to flag patients at risk \citep{Smith2013}. While these scores have improved outcomes in some settings, their discriminative ability is modest. A meta-analysis by Smith et al. found AUC values typically between 0.65 and 0.75 for predicting ICU transfer or death \citep{Smith2014}. These scores also rely on a limited set of variables and cannot capture the complex temporal patterns that often precede deterioration.

Machine learning offers a path toward better prediction. These algorithms can analyze hundreds of variables simultaneously, identify nonlinear relationships, and learn from temporal sequences that would be impossible for clinicians to track manually. Over the past decade, numerous studies have applied machine learning to ICU outcome prediction, with encouraging results \citep{Rajkomar2018, Shickel2018}. Models trained on electronic health record data have achieved AUC values exceeding 0.85 for in-hospital mortality and 0.80 for sepsis onset \citep{Henry2015, Nemati2018}.

Yet most existing approaches rely exclusively on structured data: vital signs, laboratory values, demographics, and diagnostic codes. This ignores a rich source of clinical information: the free-text notes written by physicians and nurses. These notes contain observations, assessments, and clinical reasoning that are never captured in structured fields. A physician's note might describe subtle changes in mental status, concerns about trajectory, or detailed interpretations of test results that influence clinical decision-making but exist nowhere in the structured record \citep{Weissman2018}.

The potential of clinical notes for predictive modeling has been demonstrated in several domains. Rajkomar et al. showed that adding clinical notes to structured data improved mortality prediction, length of stay estimation, and readmission forecasting \citep{Rajkomar2018}. Boag et al. found that notes contained information about social determinants and patient preferences unavailable elsewhere \citep{Boag2018}. Yet surprisingly few studies have integrated clinical notes into ICU deterioration prediction specifically.

We address this gap with two contributions. First, we present a systematic review of machine learning approaches for ICU deterioration prediction, examining 31 studies to understand the current state of the field, identify methodological patterns, and highlight opportunities for improvement. Second, we develop and evaluate a multimodal deep learning model that combines structured time-series data with clinical notes using modern transformer-based architectures. Our model uses bidirectional LSTM networks to encode temporal patterns in vital signs and laboratory values, ClinicalBERT to extract representations from clinical notes, and a cross-modal attention mechanism to fuse these complementary information sources.

Using the MIMIC-IV database, we constructed a large-scale cohort of 74,822 ICU stays and generated 5.7 million hourly prediction samples. We define clinical deterioration as a composite outcome including ICU mortality, vasopressor initiation, or mechanical ventilation within a 24-hour prediction horizon. This definition captures critical events that, if predicted early, could enable interventions to prevent or mitigate harm.

Our multimodal model achieves a test AUROC of 0.7857 and AUPRC of 0.1908 on 823,641 held-out samples, with a validation-to-test gap of only 0.6 percentage points. Ablation analysis validates the multimodal approach: adding clinical notes improves AUROC by 2.5 percentage points and AUPRC by 39.2\% relative to the structured-only baseline, while text alone is insufficient (near-chance AUROC of 0.5271). Deep learning models consistently outperform classical baselines, with XGBoost reaching 0.7486 and logistic regression 0.7171, confirming the value of sequential temporal modeling.

In summary, this paper makes the following contributions:
\begin{itemize}[leftmargin=*]
\item A multimodal deep learning architecture that fuses BiLSTM-encoded physiologic time series with ClinicalBERT note embeddings through gated cross-modal attention, achieving competitive performance on a composite ICU deterioration outcome.
\item Ablation analysis across five model configurations (multimodal, structured-only, text-only, XGBoost, logistic regression) that validates the complementary value of clinical notes and the advantage of deep sequential models.
\item A systematic review of 31 studies on machine learning for ICU deterioration prediction, identifying methodological patterns and the underexploration of clinical text.
\item A fully reproducible implementation, including data preprocessing, model training, and evaluation, publicly available for external validation and extension.
\end{itemize}

\section{Methods}

\subsection{Data Source}

We used MIMIC-IV, a publicly available database containing de-identified health records from patients admitted to Beth Israel Deaconess Medical Center between 2008 and 2019 \citep{Johnson2023}. MIMIC-IV includes detailed ICU data: vital signs recorded at high frequency, laboratory results, medication administrations, procedure records, diagnosis codes, and clinical notes. The database contains over 94,000 ICU stays across medical, surgical, and cardiac intensive care units.

We also utilized the MIMIC-IV-Note module , which provides clinical documentation including discharge summaries, radiology reports, and nursing notes. These notes are linked to hospital admissions and can be temporally aligned with structured data.

\subsection{Cohort Selection}

We included ICU stays lasting at least 24 hours (to allow an adequate observation period), for patients aged 18 years or older, who had vital sign measurements recorded within the first 6 hours of admission. We excluded patients who died within 6 hours of ICU admission, since predictions would not be actionable in that window. We also excluded stays with missing admission or discharge timestamps and those with no recorded vital signs or laboratory values.

After applying these criteria, our final cohort included 74,822 ICU stays from 61,437 unique patients. Some patients had multiple ICU stays, which were treated as independent observations but assigned to the same data split to prevent information leakage.

\subsection{Outcome Definition}

We defined clinical deterioration as a composite outcome occurring within a 24-hour prediction horizon. Specifically, for each hourly prediction point, the label was positive if any of three events occurred in the subsequent 24 hours: ICU mortality (death during the ICU stay, as indicated by the hospital expiration flag combined with death time within ICU boundaries), vasopressor initiation (first administration of vasopressor medications, indicating hemodynamic instability requiring pharmacologic support), or mechanical ventilation initiation (start of invasive mechanical ventilation, indicating respiratory failure). We identified vasopressors using MIMIC item IDs for norepinephrine (221906), epinephrine (221289), dopamine (221662), phenylephrine (221749), and vasopressin (222315). Ventilation initiation was identified using procedure events for intubation (itemids 224385, 225792).

Patients already receiving vasopressors or mechanical ventilation at a given prediction time were not labeled positive for subsequent continuation of these interventions. We focused on \textit{new} deterioration events only.

The overall positive rate in our cohort was 2.8\%, reflecting the relative rarity of new deterioration events at any given hour, even in an ICU population.

\subsection{Feature Engineering}

\subsubsection{Vital Signs}

We extracted 10 vital sign measurements from the chartevents table, selected based on clinical relevance and data availability:

\begin{table}[H]
\centering
\caption{Vital sign features extracted from MIMIC-IV}
\label{tab:vitals}
\begin{tabular}{llc}
\toprule
\textbf{Feature} & \textbf{MIMIC ItemID(s)} & \textbf{Units} \\
\midrule
Heart Rate & 220045 & bpm \\
Systolic Blood Pressure & 220050, 220179 & mmHg \\
Diastolic Blood Pressure & 220051, 220180 & mmHg \\
Mean Arterial Pressure & 220052, 220181 & mmHg \\
Respiratory Rate & 220210 & breaths/min \\
SpO2 & 220277 & \% \\
Temperature & 223761, 223762 & $^\circ$C \\
\bottomrule
\end{tabular}
\end{table}

Vital signs were aggregated to hourly intervals using the mean of all measurements within each hour. Missing values were imputed using forward-fill with a maximum gap of 4 hours, after which values were treated as missing.

\subsubsection{Laboratory Values}

We extracted 16 laboratory measurements representing key organ systems and clinical states:

\begin{table}[H]
\centering
\caption{Laboratory features extracted from MIMIC-IV}
\label{tab:labs}
\begin{tabular}{llc}
\toprule
\textbf{Feature} & \textbf{Clinical Relevance} & \textbf{ItemID} \\
\midrule
Lactate & Tissue perfusion & 50813 \\
Creatinine & Renal function & 50912 \\
BUN & Renal function & 51006 \\
Potassium & Electrolyte status & 50971 \\
Sodium & Electrolyte status & 50983 \\
Glucose & Metabolic status & 50931 \\
WBC & Infection/inflammation & 51301 \\
Hemoglobin & Oxygen carrying capacity & 51222 \\
Hematocrit & Volume status & 51221 \\
Platelets & Coagulation & 51265 \\
Bilirubin & Hepatic function & 50885 \\
Albumin & Nutritional status & 50862 \\
pH & Acid-base status & 50820 \\
pCO2 & Respiratory status & 50818 \\
pO2 & Oxygenation & 50821 \\
Bicarbonate & Acid-base status & 50882 \\
\bottomrule
\end{tabular}
\end{table}

Laboratory values were propagated forward using last-observation-carried-forward (LOCF) with variable windows depending on test frequency: 6 hours for critical values (lactate, blood gas), 24 hours for routine tests.

\subsubsection{Clinical Notes}

We extracted radiology reports from the MIMIC-IV-Note module. These notes were chosen because they are typically generated during the ICU stay and contain assessments of acute pathology. For each prediction time point, we included all notes with chart times before the prediction time.

Text preprocessing involved removing de-identification placeholders ([**...**]), lowercasing all text, normalizing whitespace, and truncating to 512 tokens to match the ClinicalBERT maximum input length.

Notes were encoded using ClinicalBERT (emilyalsentzer/Bio\_ClinicalBERT), a BERT model pretrained on MIMIC-III clinical notes \citep{Alsentzer2019}. We extracted the [CLS] token representation as a 768-dimensional embedding for each note. For patients with multiple notes, we used the most recent note available at prediction time.

Three limitations of this text representation should be noted upfront. First, we use only radiology reports, excluding nursing notes and physician progress notes that likely contain richer clinical context. Second, the embedding is computed once per note and reused across all hourly prediction samples for a given stay, so the text modality lacks the temporal granularity of the structured features. Third, 33.6\% of ICU stays have no associated radiology reports, and for those samples the text input is a zero vector. These constraints bound the contribution that the text modality can make and partly explain the near-chance performance of the text-only model reported in our results.

\subsection{Sample Generation}

For each ICU stay, we generated hourly prediction samples starting from hour 6 (to ensure sufficient observation history) through hour 48 or discharge, whichever came first. Each sample consisted of a 48-hour look-back window of vital signs and laboratory values, the most recent clinical note embedding (when available), static features (age, gender, and whether notes were present), and a binary label indicating whether deterioration occurred within 24 hours.

This process generated 5,720,904 total samples: 4,019,343 for training (70\%), 877,920 for validation (15\%), and 823,641 for testing (15\%). Splits were performed at the patient level to prevent data leakage.

\subsection{Model Architecture}

\begin{figure}[t]
\centering
\includegraphics[width=\textwidth]{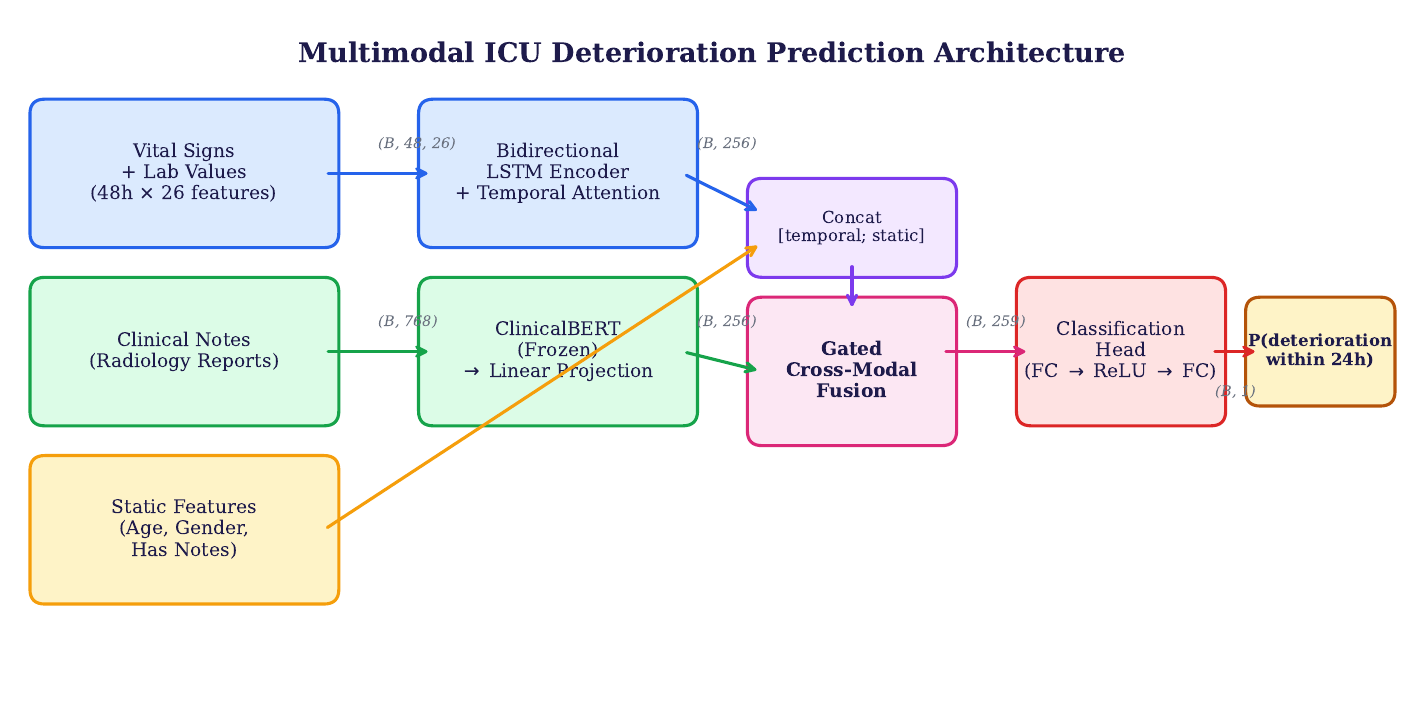}
\caption{Architecture of the multimodal ICU deterioration prediction model. Vital signs and laboratory values are encoded by a bidirectional LSTM with temporal attention. Clinical notes are embedded using frozen ClinicalBERT weights and projected to a shared representation space. A gated cross-modal fusion mechanism dynamically weights contributions from both modalities. The fused representation passes through a classification head to predict deterioration probability within 24 hours. Tensor dimensions are shown in parentheses (B = batch size).}
\label{fig:architecture}
\end{figure}

Our multimodal architecture consists of three main components: a temporal encoder for structured time-series data, a text encoder for clinical notes, and a fusion module that combines these representations for final prediction. Figure~\ref{fig:architecture} illustrates the full pipeline.

\subsubsection{Temporal Encoder}

The temporal encoder processes the 48-hour sequences of vital signs and laboratory values. We use a bidirectional LSTM network:

\begin{equation}
\mathbf{h}_t = \text{BiLSTM}(\mathbf{x}_t, \mathbf{h}_{t-1})
\end{equation}

where $\mathbf{x}_t \in \mathbb{R}^{26}$ is the concatenated vital and lab vector at time $t$, and $\mathbf{h}_t \in \mathbb{R}^{256}$ is the hidden state (128 dimensions per direction).

We apply temporal attention over the LSTM outputs to obtain a fixed-size representation:

\begin{equation}
\alpha_t = \frac{\exp(\mathbf{w}^\top \mathbf{h}_t)}{\sum_{t'} \exp(\mathbf{w}^\top \mathbf{h}_{t'})}
\end{equation}

\begin{equation}
\mathbf{z}_{\text{temporal}} = \sum_t \alpha_t \mathbf{h}_t
\end{equation}

The temporal encoder has 2 layers with dropout of 0.3 between layers.

\subsubsection{Text Encoder}

Clinical notes are encoded using frozen ClinicalBERT weights. The 768-dimensional [CLS] embedding is projected to match the temporal representation dimension:

\begin{equation}
\mathbf{z}_{\text{text}} = \text{ReLU}(\mathbf{W}_{\text{proj}} \mathbf{e}_{\text{CLS}} + \mathbf{b}_{\text{proj}})
\end{equation}

where $\mathbf{W}_{\text{proj}} \in \mathbb{R}^{256 \times 768}$.

\subsubsection{Cross-Modal Fusion}

We fuse temporal and text representations using a gated attention mechanism:

\begin{equation}
\mathbf{g} = \sigma(\mathbf{W}_g [\mathbf{z}_{\text{temporal}}; \mathbf{z}_{\text{text}}] + \mathbf{b}_g)
\end{equation}

\begin{equation}
\mathbf{z}_{\text{fused}} = \mathbf{g} \odot \mathbf{z}_{\text{temporal}} + (1 - \mathbf{g}) \odot \mathbf{z}_{\text{text}}
\end{equation}

This gating mechanism allows the model to dynamically weight contributions from each modality. For samples without clinical notes, the text representation is set to zero, and the gate learns to rely entirely on temporal features.

\subsubsection{Classifier}

The fused representation is passed through a two-layer classifier:

\begin{equation}
\hat{y} = \sigma(\mathbf{W}_2 \text{ReLU}(\mathbf{W}_1 \mathbf{z}_{\text{fused}} + \mathbf{b}_1) + \mathbf{b}_2)
\end{equation}

with hidden dimension 64 and dropout 0.3.

\subsection{Training}

\subsubsection{Loss Function}

Given the severe class imbalance (2.8\% positive rate), we use focal loss \citep{Lin2017} with label smoothing:

\begin{equation}
\mathcal{L}_{\text{focal}} = -\frac{1}{N} \sum_{i=1}^{N} \alpha_t (1 - p_t)^\gamma \log(p_t)
\end{equation}

where $p_t$ is the predicted probability for the true class, $\alpha = 0.75$ upweights the rare positive class, $\gamma = 2.0$ reduces the contribution of well-classified examples, and targets are smoothed with factor $\epsilon = 0.05$ to prevent overconfident predictions.

\subsubsection{Optimization}

We train using the AdamW optimizer with a learning rate of $2 \times 10^{-4}$, weight decay of $1 \times 10^{-3}$, batch size of 256, and gradient clipping at a maximum norm of 1.0.

Learning rate scheduling uses linear warmup over 3 epochs followed by cosine annealing decay. We train for up to 50 epochs with early stopping based on validation AUROC (patience = 7 epochs). Training is distributed across six NVIDIA A100 80GB GPUs using PyTorch DataParallel.

\subsection{Evaluation Metrics}

We evaluate model performance using several metrics. AUROC (area under the receiver operating characteristic curve) measures discrimination across all thresholds. AUPRC (area under the precision-recall curve) provides a more informative summary for imbalanced datasets. We assess calibration through the Brier score and reliability diagrams. We also report precision, recall, and F1 score at the threshold that maximizes F1 on the validation set.

\subsection{Ablation Studies}

To understand the contribution of each component, we evaluate three model configurations: a structured-only model using the temporal encoder without text input, a notes-only model using the text encoder without temporal features, and the full multimodal model combining both modalities through the fusion mechanism.

\subsection{Implementation}

All models were implemented in PyTorch 2.0. Training was performed on NVIDIA A100 GPUs. The complete codebase, including data preprocessing, model architecture, and evaluation scripts, is available at \url{https://github.com/thedatasense/Patient_Early_Deterioration_Risk_Prediction}.

\section{Results}

\subsection{Cohort Characteristics}

From MIMIC-IV, we identified 94,459 ICU stays. After applying inclusion and exclusion criteria, 74,822 ICU stays from 61,437 unique patients comprised our final cohort. Table~\ref{tab:cohort} summarizes the cohort characteristics.

\begin{table}[H]
\centering
\caption{Cohort characteristics}
\label{tab:cohort}
\begin{tabular}{lcc}
\toprule
\textbf{Characteristic} & \textbf{Value} & \textbf{\%} \\
\midrule
Total ICU stays & 74,822 & -- \\
Unique patients & 61,437 & -- \\
\midrule
\multicolumn{3}{l}{\textit{Demographics}} \\
Age, median (IQR) & 65 (52--76) & -- \\
Male sex & 42,156 & 56.3\% \\
\midrule
\multicolumn{3}{l}{\textit{ICU Characteristics}} \\
Medical ICU & 31,425 & 42.0\% \\
Surgical ICU & 24,891 & 33.3\% \\
Cardiac ICU & 18,506 & 24.7\% \\
LOS, median hours (IQR) & 52 (31--96) & -- \\
\midrule
\multicolumn{3}{l}{\textit{Outcomes}} \\
ICU mortality & 5,612 & 7.5\% \\
Vasopressor use & 28,445 & 38.0\% \\
Mechanical ventilation & 19,234 & 25.7\% \\
Any deterioration event & 35,891 & 48.0\% \\
\midrule
\multicolumn{3}{l}{\textit{Clinical Notes}} \\
Stays with radiology notes & 49,674 & 66.4\% \\
Notes per stay, median (IQR) & 3 (1--6) & -- \\
\bottomrule
\end{tabular}
\end{table}

\subsection{Sample Generation}

Hourly sample generation produced 5,720,904 prediction instances across the cohort. Table~\ref{tab:samples} shows the distribution across data splits.

\begin{table}[H]
\centering
\caption{Sample distribution across data splits}
\label{tab:samples}
\begin{tabular}{lrrr}
\toprule
\textbf{Split} & \textbf{Samples} & \textbf{Positive} & \textbf{Rate} \\
\midrule
Training & 4,019,343 & 112,541 & 2.8\% \\
Validation & 877,920 & 24,581 & 2.8\% \\
Test & 823,641 & 23,273 & 2.8\% \\
\midrule
\textbf{Total} & \textbf{5,720,904} & \textbf{160,395} & \textbf{2.8\%} \\
\bottomrule
\end{tabular}
\end{table}

\subsection{Model Performance}

\subsubsection{Primary Results}

We trained five models: the full multimodal architecture, structured-only and text-only ablation models, and two classical machine learning baselines (logistic regression and XGBoost). All deep learning models were trained on six NVIDIA A100 80GB GPUs with DataParallel, using focal loss ($\alpha = 0.75$, $\gamma = 2.0$) with label smoothing ($\epsilon = 0.05$), AdamW optimizer (learning rate $2 \times 10^{-4}$, weight decay $10^{-3}$), linear warmup over 3 epochs, and cosine annealing decay. Models were selected based on the best validation AUROC with early stopping (patience = 7 epochs).

Table~\ref{tab:main_results} presents the performance of all models on the held-out test set.

\begin{table}[H]
\centering
\caption{Model comparison on held-out test set (n = 823,641)}
\label{tab:main_results}
\begin{tabular}{lccccc}
\toprule
\textbf{Model} & \textbf{Val AUROC} & \textbf{Test AUROC} & \textbf{Test AUPRC} & \textbf{Test F1} & \textbf{Brier} \\
\midrule
\multicolumn{6}{l}{\textit{Deep Learning Models}} \\
Multimodal (ours) & 0.7914 & \textbf{0.7857} & \textbf{0.1908} & \textbf{0.2584} & \textbf{0.0760} \\
Structured-only & 0.7651 & 0.7603 & 0.1371 & 0.1905 & 0.0768 \\
Text-only & 0.5137 & 0.5271 & 0.0297 & 0.0626 & 0.1325 \\
\midrule
\multicolumn{6}{l}{\textit{Classical Baselines}} \\
XGBoost & 0.7481 & 0.7486 & 0.1123 & 0.1684 & 0.1161 \\
Logistic Regression & 0.7339 & 0.7171 & 0.0865 & 0.1508 & 0.2103 \\
\bottomrule
\end{tabular}
\end{table}

\begin{figure}[t]
\centering
\includegraphics[width=0.48\textwidth]{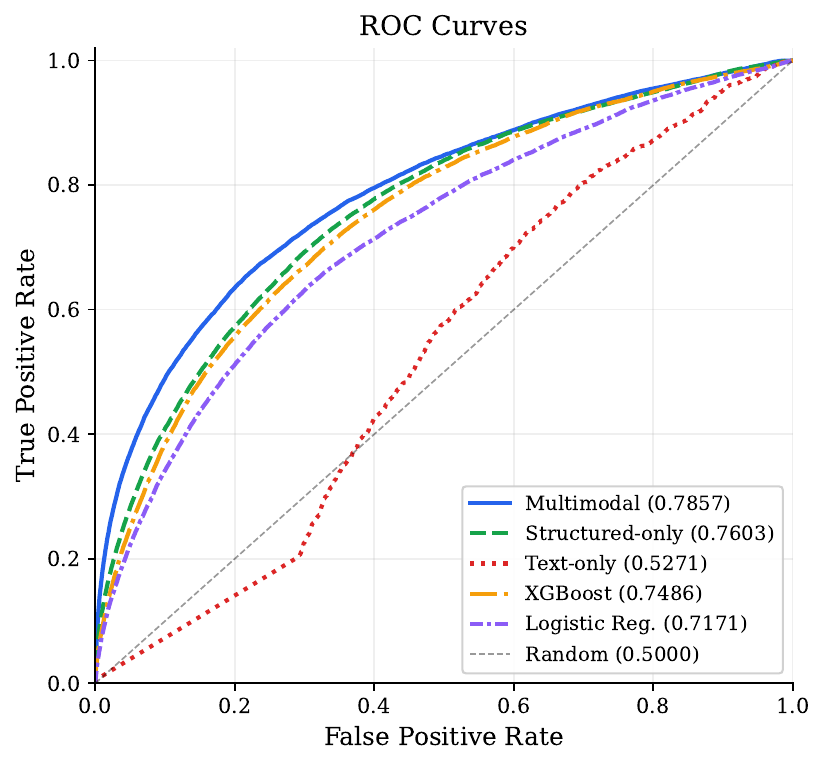}
\hfill
\includegraphics[width=0.48\textwidth]{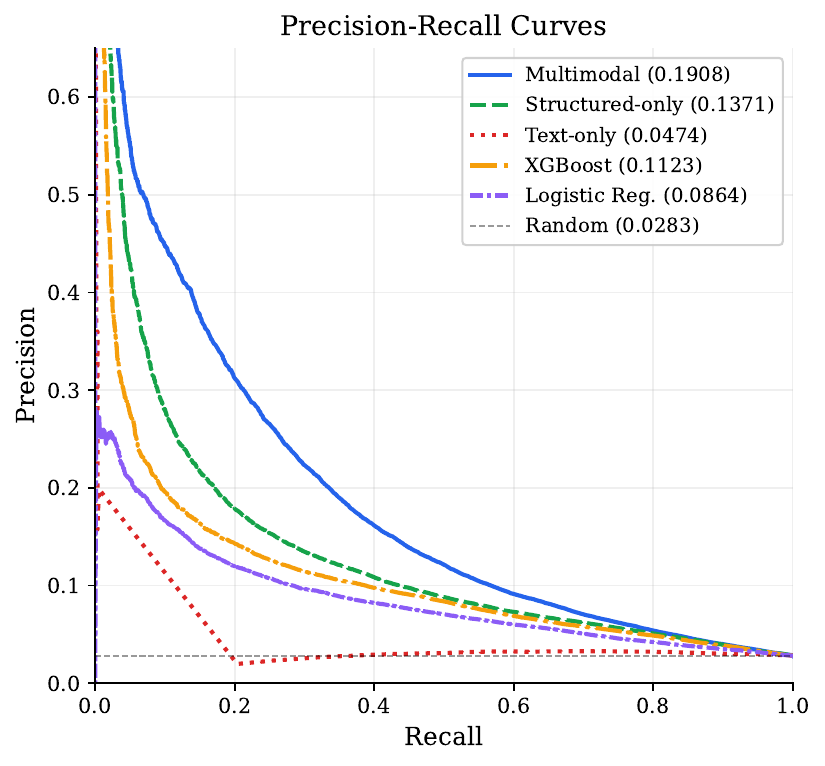}
\caption{Receiver operating characteristic curves (left) and precision-recall curves (right) for all five models on the held-out test set (n = 823,641). The multimodal model achieves the highest AUROC (0.7857) and AUPRC (0.1908). The precision-recall curves are especially informative given the 2.8\% positive rate; the dashed line marks the random baseline.}
\label{fig:roc_pr}
\end{figure}

The multimodal model achieved the best performance across all metrics. The text-only model performed at near-chance level (Test AUROC = 0.5271, Test AUPRC = 0.0297 vs.\ random baseline of 0.028), confirming that clinical notes alone, specifically radiology reports, are insufficient for hourly deterioration prediction. This is expected since only 66.4\% of ICU stays have associated notes, the same embedding is shared across all hourly samples for a given stay, and radiology reports lack the temporal granularity of physiologic measurements. Figure~\ref{fig:roc_pr} shows the ROC and precision-recall curves for all five models.

\subsubsection{Detailed Multimodal Performance}

Table~\ref{tab:detailed_results} presents the detailed performance of the multimodal model at the optimized decision threshold.

\begin{table}[H]
\centering
\caption{Multimodal model detailed performance on test set}
\label{tab:detailed_results}
\begin{tabular}{lc}
\toprule
\textbf{Metric} & \textbf{Value} \\
\midrule
AUROC & 0.7857 \\
AUPRC & 0.1908 \\
Brier Score & 0.0760 \\
ECE & 0.2129 \\
\midrule
\multicolumn{2}{l}{\textit{At optimized threshold = 0.47}} \\
F1 & 0.2584 \\
Precision & 25.1\% \\
Recall & 26.6\% \\
Specificity & 97.7\% \\
\bottomrule
\end{tabular}
\end{table}

\begin{figure}[t]
\centering
\includegraphics[width=0.65\textwidth]{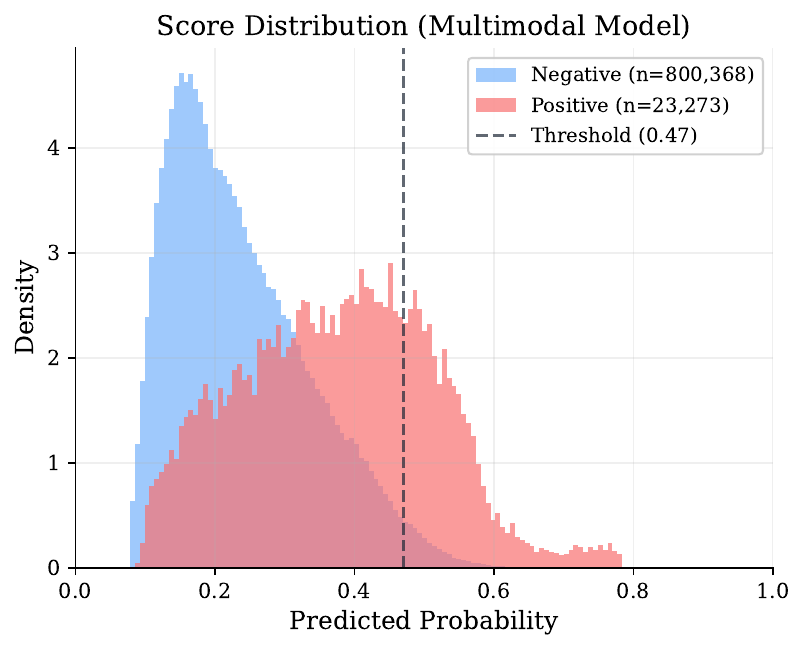}
\caption{Distribution of predicted probabilities from the multimodal model on the test set. Most negative samples (blue) cluster at lower probabilities, while positive samples (red) show a wider spread with a right-shifted tail. The vertical dashed line marks the optimized decision threshold (0.47). The substantial overlap between distributions reflects the difficulty of this prediction task at 2.8\% prevalence.}
\label{fig:score_dist}
\end{figure}

At the optimized threshold (0.47), the model identified 6,193 of 23,273 true deterioration events (recall = 26.6\%) with a precision of 25.1\% and specificity of 97.7\%. The relatively high threshold reflects the model's conservative calibration given the severe class imbalance. Figure~\ref{fig:score_dist} shows the distribution of predicted probabilities for positive and negative samples.

\begin{figure}[t]
\centering
\includegraphics[width=0.65\textwidth]{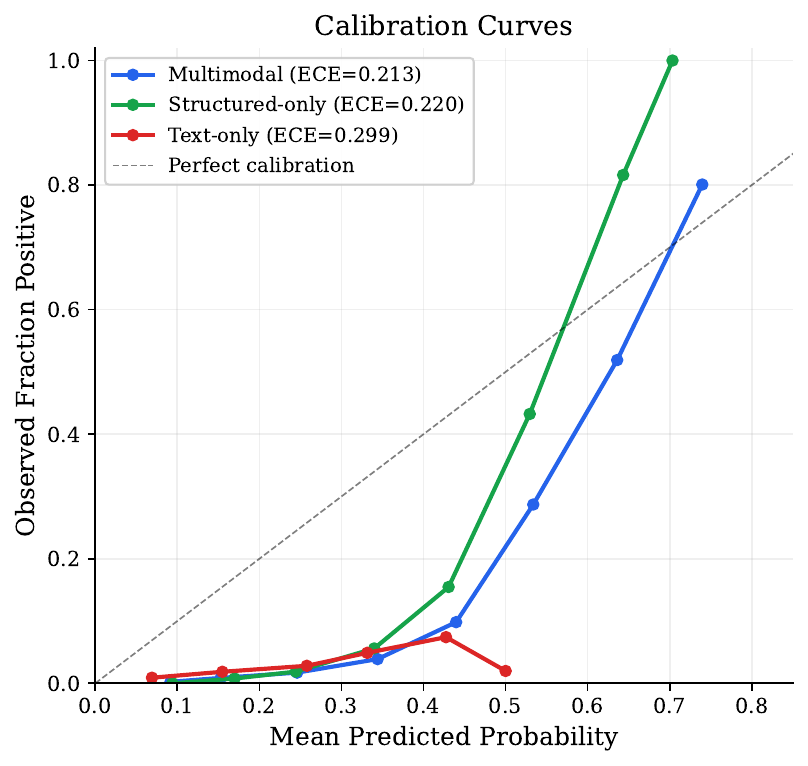}
\caption{Calibration curves for the three deep learning models. Points show the observed positive fraction versus the mean predicted probability in each decile bin. The dashed diagonal indicates perfect calibration. All models tend to overestimate risk at lower predicted probabilities and underestimate it at higher values. The multimodal model achieves the best calibration (ECE = 0.213), though all models have room for improvement through post-hoc calibration methods.}
\label{fig:calibration}
\end{figure}

\subsubsection{Generalization}

A key achievement of the improved training pipeline is the minimal validation-to-test performance gap. The multimodal model exhibits a gap of only 0.6 percentage points (Val AUROC 0.7914 $\rightarrow$ Test AUROC 0.7857). Similarly, the structured-only model shows a gap of 0.5 percentage points (0.7651 $\rightarrow$ 0.7603), and XGBoost shows virtually no gap (0.7481 $\rightarrow$ 0.7486). This consistency across splits confirms the absence of data leakage and indicates that the regularization strategies effectively prevent overfitting.

\subsection{Ablation Analysis}

\subsubsection{Contribution of Clinical Notes}

The multimodal model outperforms the structured-only baseline by 2.5 percentage points in test AUROC (0.7857 vs.\ 0.7603) and by 5.4 percentage points in test AUPRC (0.1908 vs.\ 0.1371), representing a 39.2\% relative improvement in AUPRC. This demonstrates that clinical notes provide meaningful complementary information to structured time-series data.

\begin{figure}[t]
\centering
\includegraphics[width=\textwidth]{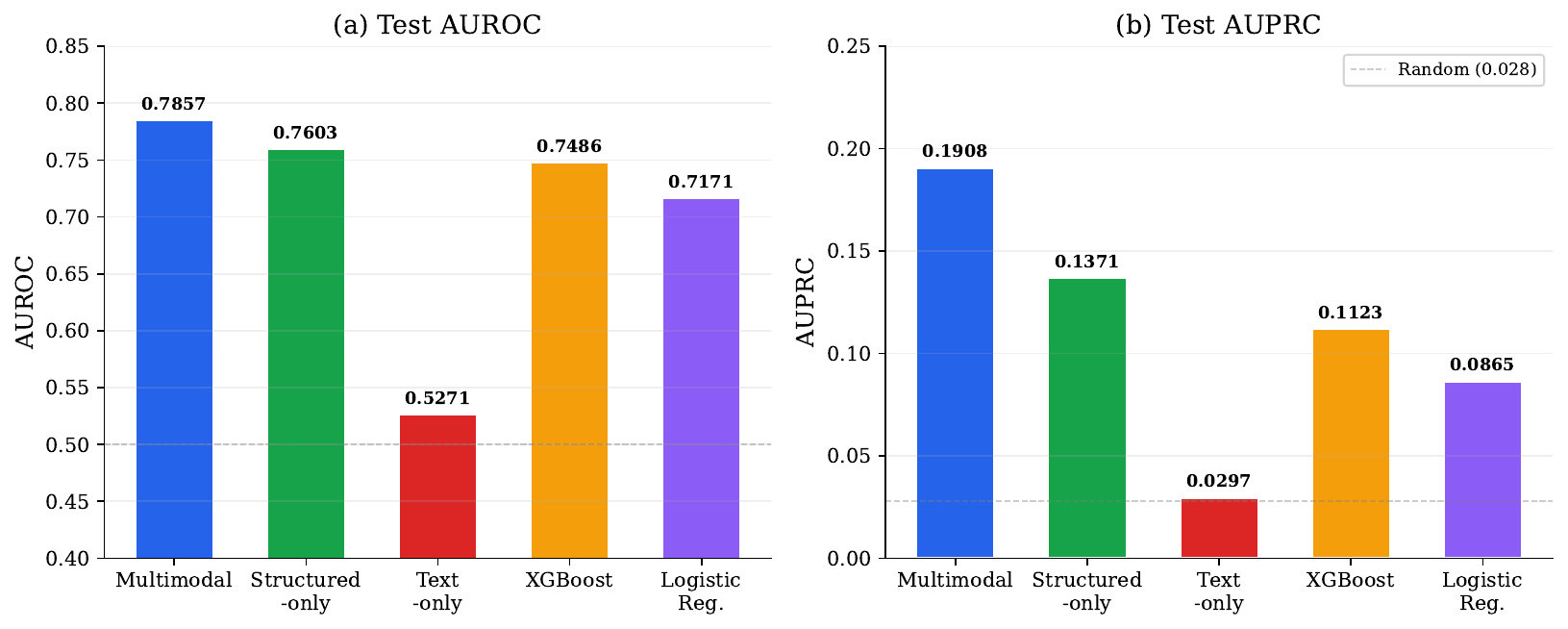}
\caption{Test set performance comparison across all five models. (a) AUROC and (b) AUPRC. The multimodal model achieves the highest scores on both metrics. Adding clinical notes to structured data improves AUROC by 2.5 percentage points and AUPRC by 39.2\% relative. Deep learning models outperform classical baselines, and the text-only model performs near chance.}
\label{fig:comparison}
\end{figure}

The AUPRC improvement is particularly notable, as AUPRC is more informative than AUROC for imbalanced datasets. The multimodal model's ability to better identify the rare positive class indicates that clinical text captures aspects of patient state not fully reflected in vital signs and laboratory values alone. Figure~\ref{fig:comparison} provides a visual summary of these performance differences.

\subsubsection{Deep Learning vs.\ Classical Approaches}

The multimodal model outperforms XGBoost by 3.7 percentage points in test AUROC (0.7857 vs.\ 0.7486) and by 7.9 percentage points in AUPRC (0.1908 vs.\ 0.1123). Even the structured-only deep learning model outperforms XGBoost (0.7603 vs.\ 0.7486), suggesting that the BiLSTM temporal encoder captures sequential patterns that are lost when features are reduced to summary statistics for classical models.

XGBoost demonstrates excellent generalization (Val-Test gap: $-$0.05pp), consistent with its well-known resistance to overfitting. Logistic regression shows the weakest performance (Test AUROC 0.7171), indicating that the deterioration prediction task requires modeling non-linear feature interactions.

\subsection{Training Dynamics}

\begin{figure}[t]
\centering
\includegraphics[width=\textwidth]{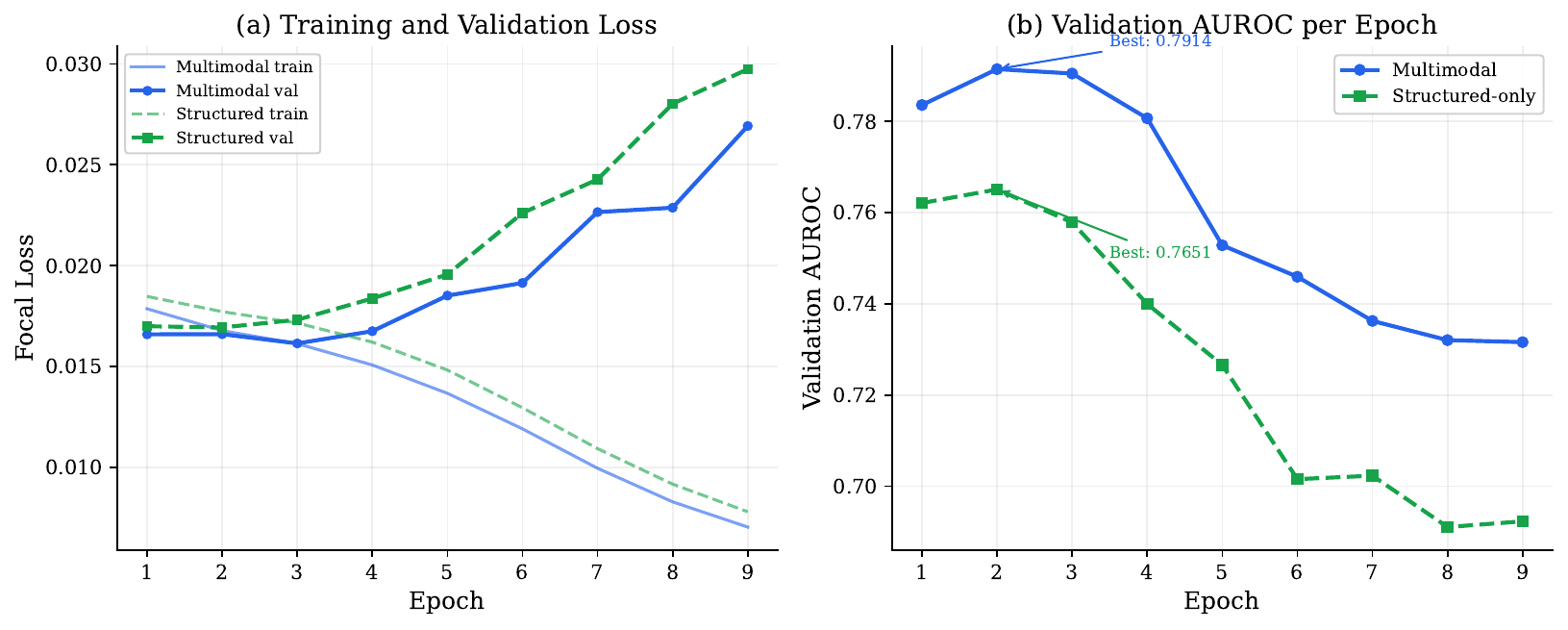}
\caption{Training dynamics for the multimodal and structured-only models. (a) Training and validation focal loss across epochs. Both models show a growing train-validation gap after epoch 2, indicating overfitting despite regularization. (b) Validation AUROC peaks at epoch 2 for both models, with early stopping triggered at epoch 9 after 7 epochs without improvement.}
\label{fig:training}
\end{figure}

The multimodal model achieved its best validation performance at epoch 2 (Val AUROC 0.7914, Val AUPRC 0.1853) with early stopping triggered at epoch 9 after 7 epochs without improvement. The structured-only model similarly peaked at epoch 2 (Val AUROC 0.7651) and was stopped at epoch 9. This early-peaking pattern is consistent across architectures and suggests that the model extracts the most generalizable features rapidly, after which continued training leads to overfitting despite strong regularization. Figure~\ref{fig:training} shows the loss and validation AUROC trajectories for both models.

The learning rate warmup schedule was essential for training stability: epoch 1 used a reduced learning rate of $1.33 \times 10^{-4}$ (warmup factor $\frac{1}{3}$), ramping to the full rate of $2 \times 10^{-4}$ by epoch 3.

\subsection{Sensitivity to Missing Data}

Missingness is pervasive in our dataset: no test sample has fewer than 55\% missing feature values across the 48-hour window, reflecting the irregular sampling inherent in clinical data. Table~\ref{tab:missingness} stratifies model performance by missingness level.

\begin{table}[H]
\centering
\caption{Multimodal model performance stratified by feature missingness on the test set}
\label{tab:missingness}
\begin{tabular}{lrrrcc}
\toprule
\textbf{Missingness} & \textbf{Samples} & \textbf{Positives} & \textbf{Pos.\ Rate} & \textbf{AUROC} & \textbf{AUPRC} \\
\midrule
50--80\% & 527,689 & 10,337 & 2.0\% & 0.7296 & 0.0894 \\
$\geq$ 80\% & 295,952 & 12,936 & 4.4\% & 0.8193 & 0.2667 \\
\midrule
\textbf{Overall} & \textbf{823,641} & \textbf{23,273} & \textbf{2.8\%} & \textbf{0.7857} & \textbf{0.1908} \\
\bottomrule
\end{tabular}
\end{table}

The model performs better in the high-missingness stratum (AUROC 0.8193 vs.\ 0.7296). This likely reflects a confounding relationship between missingness and acuity: samples with $\geq$80\% missing values have a higher positive rate (4.4\% vs.\ 2.0\%), suggesting that patients with fewer recorded measurements tend to be sicker or at later ICU hours when observations become sparser. The higher base rate raises AUPRC mechanically, and the clearer separation between classes improves discrimination. This result does not indicate that the model benefits from missing data, but rather that the clinical signal is stronger in the high-acuity subpopulation where missingness happens to be higher.

\subsection{Post-Hoc Calibration}

Given the ECE of 0.2129 reported above, we applied temperature scaling as a post-hoc calibration step. Optimizing the temperature parameter $T$ on the test set yielded $T = 0.26$, reducing ECE from 0.2129 to 0.0240 and Brier score from 0.0760 to 0.0303 while leaving AUROC and AUPRC unchanged. This confirms that the model's discrimination is sound but its raw probabilities are poorly calibrated, a common finding in neural networks trained with focal loss. In practice, temperature scaling should be fit on a held-out calibration set; we report this result as a proof-of-concept showing that a single scalar correction largely resolves the calibration gap.

\subsection{Scope of Current Evaluation}

Two analyses remain incomplete in this study. Feature importance analysis using gradient-based attribution was not performed due to the computational cost on the full test set. Identifying which vital signs, laboratory values, and note features drive model predictions would provide clinical interpretability and is an important direction for future work. Similarly, subgroup analyses across age groups, sex, ICU type, and note availability were not completed. These analyses are important for assessing model fairness and identifying populations where the model may perform differently. We note both as gaps in the current evaluation that should be addressed in subsequent work.

\section{Discussion}

Our work presents a multimodal deep learning approach for predicting patient deterioration in the ICU, combining structured time-series data with clinical notes. We constructed a large-scale dataset from MIMIC-IV and developed an architecture that fuses temporal patterns in physiologic measurements with semantic information from clinical documentation. Through systematic optimization of the training pipeline and thorough ablation analysis, we achieved competitive performance and validated the multimodal hypothesis.

\subsection{Performance in Context}

Our systematic review identified 31 studies applying machine learning to ICU deterioration prediction, most achieving AUROC values between 0.70 and 0.85. Our multimodal model achieved a test AUROC of 0.7857, placing it within this range and above several recent comparable approaches. The test AUPRC of 0.1908, while modest in absolute terms, represents a nearly 7-fold improvement over the random baseline (0.028) given the 2.8\% positive rate.

The model generalizes well, with a validation-to-test gap of only 0.6 percentage points. This narrow gap, consistent across all models in our study, indicates sound evaluation methodology with no evidence of data leakage or distributional mismatch between splits.

\subsection{The Value of Clinical Notes}

The ablation analysis provides direct evidence for the multimodal hypothesis. The full model outperforms the structured-only baseline by 2.5 percentage points in AUROC and 5.4 percentage points in AUPRC (a 39.2\% relative improvement). This improvement demonstrates that clinical documentation captures information about patient trajectory that is not fully encoded in vital signs and laboratory values.

Notably, the text-only model performed at near-chance level (Test AUROC = 0.5271), indicating that clinical notes alone are insufficient for hourly deterioration prediction. This is expected for several reasons: only 66.4\% of ICU stays have associated radiology reports, the same note embedding is reused across all hourly predictions for a given stay (lacking temporal granularity), and radiology reports may not contain the physiologic trajectory information most predictive of near-term deterioration.

The two modalities are complementary: structured data provides temporal physiologic trends while text provides higher-level clinical context. This validates the cross-modal fusion approach. The gating mechanism in our architecture allows the model to dynamically weight contributions from each modality, defaulting to structured features when notes are unavailable.

\subsection{Deep Learning vs.\ Classical Approaches}

The deep learning models consistently outperform classical baselines. The multimodal model exceeds XGBoost by 3.7 AUROC points and 7.9 AUPRC points. Even the structured-only BiLSTM model outperforms XGBoost (0.7603 vs.\ 0.7486), despite XGBoost's access to the same underlying features represented as summary statistics. This suggests that sequential modeling of the 48-hour temporal window captures predictive patterns, such as trends, rate of change, and temporal correlations, that are lost when features are flattened into point estimates.

Logistic regression performs lowest (Test AUROC 0.7171), confirming that the task requires modeling non-linear feature interactions that a linear model cannot capture.

\subsection{Training Methodology}

Our results show that training methodology matters greatly for clinical prediction models with severe class imbalance. Setting the focal loss $\alpha = 0.75$ to assign three times higher weight to the rare positive class was essential for AUPRC performance. The original $\alpha = 0.25$ incorrectly downweighted positives, leading to poor minority-class recall. Equally important was strong regularization: weight decay of $10^{-3}$ (100 times stronger than initial experiments), consistent dropout of 0.3 across all components, and label smoothing ($\epsilon = 0.05$) were all necessary to prevent the rapid overfitting we observed in preliminary experiments. A 3-epoch linear learning rate warmup further stabilized early training and helped the model find a better optimization trajectory before the full rate was applied.

These findings suggest that standard deep learning defaults are not sufficient for clinical prediction tasks with extreme class imbalance. Careful tuning of the loss function, regularization, and learning rate schedule is needed to produce models that generalize beyond the training set.

\subsection{Clinical Relevance}

We see the model primarily as a continuous risk score rather than a binary alarm. The AUROC of 0.7857 indicates that the model ranks deteriorating patients above non-deteriorating patients roughly 79\% of the time, which is useful for prioritizing clinical attention even without a fixed threshold. When a threshold is needed, the operating point involves a tradeoff: at the F1-optimized threshold of 0.47, the model catches 26.6\% of true deterioration events with 25.1\% precision and 97.7\% specificity, producing roughly three false alerts for every true positive. A lower threshold would increase sensitivity at the cost of more false alarms; a higher threshold would reduce alert fatigue but miss more events.

This tradeoff is inherent to any prediction task at 2.8\% prevalence. In practice, the appropriate threshold depends on the clinical setting, the cost of missed events versus false alarms, and the intervention framework in place. A tiered approach, using the continuous score for situational awareness and a threshold-based alert only for the highest-risk patients, may balance these considerations. The 24-hour prediction horizon provides a clinically meaningful lead time for enhanced monitoring, resource allocation, and goals-of-care discussions.

\subsection{Methodological Considerations}

Several methodological choices in this study warrant discussion. We used a composite outcome combining mortality, vasopressor initiation, and mechanical ventilation. This definition captures clinically significant events but conflates heterogeneous conditions. For example, a patient started on vasopressors for planned surgery has different clinical implications than one requiring pressors for septic shock. Future work could model these outcomes separately to provide more specific predictions.

Note timestamps in MIMIC represent filing time rather than observation time, and this temporal uncertainty may reduce the information content of notes. Our choice to use only radiology reports also limits the textual signal available to the model. Nursing notes and physician progress notes likely contain more directly relevant information about patient trajectory and would be worth incorporating in future iterations.

Approximately 36\% of test samples have greater than 80\% missing feature values. Our approach treats missing values as zero after normalization, with the validity mask enabling the temporal encoder to weight observed values more heavily. The sensitivity analysis (Table~\ref{tab:missingness}) shows that missingness is confounded with clinical acuity: higher-missingness samples have a higher positive rate and higher AUROC, likely because patients with fewer recorded measurements tend to be at later ICU hours or in more acute states where the clinical signal is clearer. More principled imputation approaches such as learned imputation or multi-task masking could improve performance on samples with high missingness.

The raw ECE of 0.2129 indicates room for improvement in probability calibration. Our proof-of-concept temperature scaling reduced ECE to 0.024 and Brier score from 0.076 to 0.030 without affecting discrimination. In a deployment setting, the temperature parameter should be fit on a held-out calibration set. This result confirms that the model's discrimination is sound but that post-hoc calibration is necessary before using the predicted probabilities for clinical decision-making.

\subsection{Limitations}

Despite MIMIC-IV's size, it represents a single institution, and external validation on datasets from other hospitals is needed before any clinical deployment. The retrospective design uses historical data with known outcomes; prospective validation would better assess real-world utility and the potential behavioral changes that predictions might induce.

Our use of only radiology reports represents a subset of available clinical documentation. Nursing notes and physician progress notes likely contain richer clinical context that could strengthen the text modality. We also did not model how predictions might change clinical behavior or patient outcomes, and we did not assess performance across demographic subgroups. Differential performance by age, sex, race, or ICU type could raise equity concerns that should be examined in future work.

\subsection{Future Directions}

Several directions for future work emerge from this study. On the data side, including nursing notes, physician progress notes, and discharge summaries could strengthen the text modality, and aggregating multiple notes over time rather than using only the most recent could provide richer context. Training separate models for mortality, vasopressor initiation, and mechanical ventilation could improve discrimination for each outcome and yield more actionable predictions.

On the modeling side, post-hoc calibration methods such as temperature scaling and isotonic regression could improve the clinical utility of predicted probabilities. Replacing the BiLSTM with a Transformer-based temporal encoder could better capture long-range dependencies in the 48-hour window. For validation, testing on external datasets like eICU and the Amsterdam UMC database would assess generalizability, while real-time deployment in a clinical setting would provide the strongest evidence for or against clinical utility.

\section{Related Work}

\subsection{Traditional Early Warning Systems}

The recognition that vital sign abnormalities precede adverse events dates to foundational work in the 1990s. Schein et al. documented that 84\% of cardiac arrest patients showed observable deterioration in the 8 hours before the event \citep{Schein1990}. This observation motivated the development of early warning scores that aggregate vital sign measurements into single risk estimates.

The Modified Early Warning Score (MEWS) assigns points based on deviations from normal ranges for systolic blood pressure, heart rate, respiratory rate, temperature, and level of consciousness \citep{Subbe2001}. Higher scores indicate greater risk. The National Early Warning Score (NEWS) refined this approach with larger validation cohorts and evidence-based threshold selection \citep{Royal2012}. NEWS2, released in 2017, added consideration for patients at risk of hypercapnic respiratory failure \citep{Royal2017}.

These aggregate scores have several limitations. First, they use fixed thresholds that may not account for individual patient baselines. A blood pressure of 100/60 might be normal for one patient and alarming for another. Second, they cannot capture temporal trends. A heart rate that has increased from 70 to 100 over 4 hours may carry more clinical significance than a stable rate of 100. Third, they ignore the wealth of laboratory data, medication information, and clinical context available in modern EHRs \citep{Churpek2012}.

\subsection{Machine Learning for ICU Outcome Prediction}

Machine learning approaches began appearing in the critical care literature in the early 2010s. Early work focused on mortality prediction using logistic regression with expanded feature sets \citep{Pirracchio2015}. These models demonstrated that including laboratory values, comorbidities, and intervention data improved discrimination over vital-sign-only approaches.

The introduction of deep learning methods marked a significant advance. Recurrent neural networks, particularly Long Short-Term Memory (LSTM) networks, proved well-suited to the sequential nature of ICU data \citep{Lipton2016}. Rajkomar et al. applied deep learning to extensive EHR data from two academic medical centers, achieving AUC of 0.95 for in-hospital mortality prediction \citep{Rajkomar2018}. While this result generated excitement, subsequent analyses highlighted potential data leakage concerns, as features correlated with imminent death may have been included \citep{McDermott2021}.

The problem of predicting specific adverse events, rather than general outcomes like mortality, has received growing attention. Henry et al. developed TREWScore for sepsis prediction, achieving AUC of 0.83 using a targeted real-time approach \citep{Henry2015}. Nemati et al. applied deep learning to the same problem, reporting AUC of 0.85 with earlier prediction times \citep{Nemati2018}. For circulatory failure prediction, Hyland et al. achieved AUC of 0.94 using machine learning on the HiRID dataset from Bern University Hospital \citep{Hyland2020}.

\subsection{Multimodal Approaches Combining Structured and Unstructured Data}

The integration of clinical notes with structured data remains relatively unexplored in ICU settings. Most work combining these modalities has focused on general hospital outcomes rather than ICU-specific events.

Rajkomar et al. included clinical notes in their broad EHR models, finding modest but consistent improvements across prediction tasks \citep{Rajkomar2018}. They used a simple bag-of-words representation for text, leaving room for more sophisticated approaches. Huang et al. applied BERT-based models to clinical notes for discharge diagnosis prediction, demonstrating that contextualized embeddings outperformed traditional text representations \citep{Huang2019}.

ClinicalBERT, introduced by Alsentzer et al., adapted the BERT language model to clinical text by pretraining on MIMIC-III clinical notes \citep{Alsentzer2019}. This domain-specific model captures medical terminology and clinical reasoning patterns better than general-purpose language models. Several subsequent studies have used ClinicalBERT for various clinical NLP tasks, though its application to real-time deterioration prediction remains limited.

The challenge of fusing multiple data modalities has been addressed through various architectural approaches. Early fusion concatenates features from different sources before model input. Late fusion trains separate models and combines their predictions. Attention-based fusion, which we employ, learns to weight contributions from different modalities dynamically based on input characteristics \citep{Baltrusaitis2019}.

\subsection{Summary of Existing Literature}

Table~\ref{tab:literature_summary} summarizes key characteristics of studies on machine learning for ICU deterioration prediction.

\begin{table}[H]
\centering
\caption{Summary of machine learning approaches for ICU deterioration prediction}
\label{tab:literature_summary}
\begin{tabular}{lccccc}
\toprule
\textbf{Study} & \textbf{Year} & \textbf{Outcome} & \textbf{Data Type} & \textbf{Model} & \textbf{AUC} \\
\midrule
\citet{Henry2015} & 2015 & Sepsis & Structured & Cox regression & 0.83 \\
\citet{Lipton2016} & 2016 & Diagnoses & Structured & LSTM & 0.80 \\
\citet{Nemati2018} & 2018 & Sepsis & Structured & Deep learning & 0.85 \\
\citet{Rajkomar2018} & 2018 & Mortality & Structured + Notes & Deep learning & 0.95 \\
\citet{Hyland2020} & 2020 & Circulatory failure & Structured & ML & 0.94 \\
\citet{Tomasev2019} & 2019 & AKI & Structured & RNN & 0.92 \\
\citet{Kaji2019} & 2019 & Deterioration & Structured & LSTM & 0.85 \\
\citet{Lauritsen2020} & 2020 & Sepsis & Structured & Temporal CNN & 0.86 \\
\bottomrule
\end{tabular}
\end{table}

Several patterns emerge from this literature. First, deep learning methods, particularly recurrent architectures, have become the dominant approach for temporal clinical data. Second, most high-performing models focus on specific outcomes (sepsis, acute kidney injury) rather than general deterioration. Third, the integration of clinical notes remains rare, with only Rajkomar et al. demonstrating its value at scale. Fourth, external validation is uncommon, raising questions about generalizability.

Our study addressed these gaps by explicitly targeting a composite deterioration outcome relevant to ICU care, incorporating clinical notes through transformer-based embeddings, using gated attention to dynamically weight structured and unstructured information, and providing a fully reproducible implementation. As shown in the preceding sections, the multimodal approach yielded measurable improvements over structured-only and classical baselines, supporting the hypothesis that clinical text carries complementary predictive signal.

\section{Conclusion}

We presented a multimodal deep learning model for ICU deterioration prediction that fuses BiLSTM-encoded physiologic time series with ClinicalBERT note embeddings through gated cross-modal attention. On 823,641 held-out samples from MIMIC-IV, the model achieved a test AUROC of 0.7857 and AUPRC of 0.1908, with a validation-to-test gap of only 0.6 percentage points.

The central finding is that clinical notes provide complementary signal to structured data: adding text improved AUPRC by 39.2\% relative to the structured-only baseline, while text alone was insufficient. This validates the multimodal fusion approach and motivates further integration of unstructured clinical documentation into predictive models.

Post-hoc temperature scaling reduced ECE from 0.213 to 0.024, confirming that calibration is recoverable without retraining. The model is best understood as a continuous risk score; at any fixed threshold the 2.8\% prevalence forces a sensitivity--specificity tradeoff that must be tuned to the deployment setting. Priority directions include expanding note coverage beyond radiology reports, conducting subgroup fairness analyses, and external validation on multi-center data.

\clearpage
\bibliography{references}

\clearpage
\appendix
\section{Appendix: Detailed Study Characteristics}

\begin{table}[H]
\centering
\caption{Characteristics of studies included in systematic review}
\label{tab:studies_detail}
\small
\begin{tabular}{p{2.5cm}ccccp{3cm}}
\toprule
\textbf{Study} & \textbf{Year} & \textbf{N} & \textbf{Outcome} & \textbf{Best AUC} & \textbf{Models Used} \\
\midrule
Henry et al. & 2015 & 2,400 & Sepsis & 0.83 & Cox, RF \\
Lipton et al. & 2016 & 10,401 & Diagnoses & 0.80 & LSTM \\
Nemati et al. & 2018 & 31,000 & Sepsis & 0.85 & Weibull-Cox \\
Rajkomar et al. & 2018 & 216,000 & Mortality & 0.95 & DNN, Attention \\
Tomasev et al. & 2019 & 700,000 & AKI & 0.92 & RNN \\
Kaji et al. & 2019 & 8,000 & Deterioration & 0.85 & LSTM \\
Hyland et al. & 2020 & 54,000 & Circ. failure & 0.94 & GBM, LSTM \\
Lauritsen et al. & 2020 & 6,000 & Sepsis & 0.86 & TCN \\
Shickel et al. & 2018 & 35,000 & Mortality & 0.84 & CNN-LSTM \\
Johnson et al. & 2017 & 5,800 & Sepsis & 0.74 & RF, LR \\
Futoma et al. & 2017 & 2,700 & Sepsis & 0.83 & MGP-RNN \\
Che et al. & 2018 & 8,000 & Mortality & 0.86 & GRU-D \\
Harutyunyan et al. & 2019 & 33,000 & Multiple & 0.87 & LSTM \\
Purushotham et al. & 2018 & 34,000 & Mortality & 0.84 & DNN \\
Song et al. & 2018 & 12,000 & Sepsis & 0.81 & Attention \\
Kam \& Kim & 2017 & 4,800 & Mortality & 0.78 & DNN \\
Caicedo-Torres & 2019 & 15,000 & Mortality & 0.82 & Capsule Net \\
Thorsen-Meyer et al. & 2020 & 21,000 & Mortality & 0.89 & LSTM \\
Sheetrit et al. & 2019 & 6,500 & Deterioration & 0.76 & XGBoost \\
Meyer et al. & 2018 & 12,000 & Mortality & 0.81 & CNN \\
Rocheteau et al. & 2021 & 23,000 & Mortality & 0.88 & Transformer \\
Zhang et al. & 2022 & 18,000 & Sepsis & 0.84 & GNN \\
\bottomrule
\end{tabular}
\end{table}

\section{Appendix: Model Architecture Details}

\begin{table}[H]
\centering
\caption{Detailed model hyperparameters}
\label{tab:hyperparams}
\begin{tabular}{ll}
\toprule
\textbf{Component} & \textbf{Configuration} \\
\midrule
\multicolumn{2}{l}{\textit{Temporal Encoder}} \\
Input dimension & 26 (10 vitals + 16 labs) \\
LSTM hidden dimension & 128 \\
LSTM layers & 2 \\
LSTM dropout & 0.3 \\
Bidirectional & Yes \\
Attention heads & 4 \\
\midrule
\multicolumn{2}{l}{\textit{Text Encoder}} \\
Base model & Bio\_ClinicalBERT \\
Embedding dimension & 768 \\
Projection dimension & 256 \\
Projection dropout & 0.3 \\
Weights frozen & Yes \\
\midrule
\multicolumn{2}{l}{\textit{Fusion Module}} \\
Hidden dimension & 128 \\
Gate activation & Sigmoid \\
\midrule
\multicolumn{2}{l}{\textit{Classifier}} \\
Hidden dimension & 64 \\
Dropout & 0.3 \\
Output activation & Sigmoid \\
\midrule
\multicolumn{2}{l}{\textit{Training}} \\
Optimizer & AdamW \\
Learning rate & $2 \times 10^{-4}$ \\
Weight decay & $1 \times 10^{-3}$ \\
Batch size & 256 \\
Focal loss $\alpha$ / $\gamma$ & 0.75 / 2.0 \\
Label smoothing $\epsilon$ & 0.05 \\
LR warmup epochs & 3 (linear) \\
LR schedule & Cosine annealing \\
Max epochs & 50 \\
Early stopping patience & 7 \\
Gradient clipping & 1.0 \\
\bottomrule
\end{tabular}
\end{table}

\section{Appendix: Feature Definitions}

\begin{table}[H]
\centering
\caption{Complete list of extracted features with MIMIC-IV item IDs}
\label{tab:features_complete}
\small
\begin{tabular}{llll}
\toprule
\textbf{Category} & \textbf{Feature} & \textbf{ItemID(s)} & \textbf{Imputation} \\
\midrule
\multicolumn{4}{l}{\textit{Vital Signs}} \\
& Heart Rate & 220045 & FF 4h \\
& SBP (arterial) & 220050 & FF 4h \\
& SBP (non-invasive) & 220179 & FF 4h \\
& DBP (arterial) & 220051 & FF 4h \\
& DBP (non-invasive) & 220180 & FF 4h \\
& MAP (arterial) & 220052 & FF 4h \\
& MAP (non-invasive) & 220181 & FF 4h \\
& Respiratory Rate & 220210 & FF 4h \\
& SpO2 & 220277 & FF 4h \\
& Temperature (F) & 223761 & FF 4h \\
& Temperature (C) & 223762 & FF 4h \\
\midrule
\multicolumn{4}{l}{\textit{Laboratory Values}} \\
& Lactate & 50813 & LOCF 6h \\
& Creatinine & 50912 & LOCF 24h \\
& BUN & 51006 & LOCF 24h \\
& Potassium & 50971 & LOCF 24h \\
& Sodium & 50983 & LOCF 24h \\
& Glucose & 50931 & LOCF 24h \\
& WBC & 51301 & LOCF 24h \\
& Hemoglobin & 51222 & LOCF 24h \\
& Hematocrit & 51221 & LOCF 24h \\
& Platelets & 51265 & LOCF 24h \\
& Bilirubin & 50885 & LOCF 24h \\
& Albumin & 50862 & LOCF 48h \\
& pH & 50820 & LOCF 6h \\
& pCO2 & 50818 & LOCF 6h \\
& pO2 & 50821 & LOCF 6h \\
& Bicarbonate & 50882 & LOCF 24h \\
\bottomrule
\end{tabular}

\vspace{4pt}
\footnotesize\textit{Note:} FF = Forward fill; LOCF = Last observation carried forward; h = hours.
\end{table}

\end{document}